\pdfoutput=1

\documentclass[11pt]{article}

\usepackage[final]{acl2025}

\usepackage{times}
\usepackage{latexsym}
\usepackage{amsmath}
\usepackage{amsfonts}
\usepackage{multirow}
\usepackage{subcaption}
\usepackage{caption}
\usepackage{booktabs}
\usepackage[utf8]{inputenc}
\usepackage[T1]{fontenc}

\usepackage{CJKutf8}
\usepackage{afterpage} 

\usepackage{microtype}

\usepackage{inconsolata}

\usepackage{graphicx}

\DeclareMathOperator*{\argmax}{arg\,max}

\title {HENT-SRT: Hierarchical Efficient Neural Transducer with Self-Distillation for Joint Speech Recognition and Translation}

\author{
Amir Hussein \\
CLSP \\
JHU \\
\And
Cihan Xiao \\
CLSP\\
JHU \\
\And
Matthew Wiesner \\
HLTCOE \\
JHU \\
\AND
Dan Povey \\
Xiaomi \\ 
\And
Leibny Paola Garcia \\
HLTCOE/CLSP\\
JHU \\
\And
Sanjeev Khudanpur \\
HLTCOE/CLSP \\
JHU
}

\begin{document}
\maketitle
\begin{abstract}
Neural transducers (NT) provide an effective framework for speech streaming, demonstrating strong performance in automatic speech recognition (ASR). However, the application of NT to speech translation (ST) remains challenging, as existing approaches struggle with word reordering and performance degradation when jointly modeling ASR and ST, resulting in a gap with attention-based encoder-decoder (AED) models. Existing NT-based ST approaches also suffer from high computational training costs. To address these issues, we propose HENT-SRT (Hierarchical Efficient Neural Transducer for Speech Recognition and Translation), a novel framework that factorizes ASR and translation tasks to better handle reordering. To ensure robust ST while preserving ASR performance, we use self-distillation with CTC consistency regularization. Moreover, we improve computational efficiency by incorporating best practices from ASR transducers, including a down-sampled hierarchical encoder, a stateless predictor, and a pruned transducer loss to reduce training complexity. Finally, we introduce a blank penalty during decoding, reducing deletions and improving translation quality. Our approach is evaluated on three conversational datasets Arabic, Spanish, and Mandarin achieving new state-of-the-art performance among NT models and substantially narrowing the gap with AED-based systems.
\end{abstract}

\begingroup
\makeatletter
\renewcommand{\footnoterule}{}
\renewcommand\@makefnmark{}
\renewcommand{\@makefntext}[1]{\noindent#1}
\makeatother
\footnotetext[0]{Accepted to IWSLT 2025.}

\section{Introduction}
Translation of spoken conversations across languages plays a crucial role in cross-cultural communication, healthcare, 
and education \cite{koksal2020role, nakamura2009overcoming, al2020implications}. Traditionally, speech translation (ST) systems have been built using a cascaded approach, where automatic speech recognition (ASR) first transcribes speech into text, which is then passed to a machine translation (MT) system \citep{matusov2005integration, bertoldi2005new, sperber2017toward, pino2019harnessing, yang2022jhu}. This modular approach facilitates the use of large text corpora for MT training, at the cost of (1) complex beam search algorithms in streaming applications \citep{rabatin2024navigating}, (2) error propagation from the ASR to MT model, (3) an inability to leverage paralinguistic information such as prosody, and (4) additional latency, as MT processing must wait for ASR to complete.  

To overcome the limitations of cascaded systems, end-to-end speech translation (E2E-ST) has emerged as a promising approach that directly maps source speech to target text, providing a more streamlined architecture, reduced latency, and competitive performance~\citep{berard2016listen, berard2018endtoend, dalmia2021searchable, gaido2020end, yan-etal-2023-espnet}. However, most end-to-end speech translation (E2E-ST) research has focused on offline attention-based encoder-decoder (AED) architectures. As label-synchronous systems, AEDs require multiple input frames before emitting each output token, which limits their suitability for streaming applications and increases sensitivity to utterance segmentation \citep{anastasopoulos2022findings,sinclair14_interspeech}. To enable streaming in AED models, researchers have proposed wait-k policies, which introduce a controlled buffering mechanism to balance latency and translation quality \citep{ma-etal-2020-simulmt, chen-etal-2021-direct, ma2021streaming}. However, finding the optimal wait-k policy is challenging,  as it must balance latency and quality, and the necessary buffering cause delays, making AED-based models less suitable for streaming applications.

In contrast, frame-synchronous architectures like Connectionist Temporal Classification (CTC) \citep{graves2006connectionist} and the Neural Transducer (NT) \citep{graves2012sequence} more naturally handle streaming data and demonstrate greater robustness to utterance segmentation, mitigating the over- and under-generation issues in AED models \citep{chiu2019comparison,yan2023ctc}. While CTC enforces strictly monotonic alignments, which limits its ability to handle word reordering \citep{yan2023ctc}, the neural transducer (NT) relaxes these constraints, allowing the generation of longer output sequences and the modeling of autoregressive token dependencies. This makes NT more suitable for translation, as illustrated in Figure~\ref{fig:transducer}. To improve NT's reordering capabilities, researchers have proposed augmenting the joiner with cross-attention \citep{liu2021cross} and the predictor with an AED-based decoder \citep{tang2023hybrid}, though these additions increase computational complexity and latency. To further enhance translation quality, \citep{tang2023hybrid} introduced attention pooling for better encoder-predictor fusion at the frame level, while \citep{xue2022large} explored similar mechanisms. Additionally, \citep{wang23oa_interspeech} proposed a transducer-based model for unified ASR and ST, but its shared encoder struggles with reordering, making it less effective than AED in offline settings. 

\begin{figure*}[!tb]
  \centering
  \begin{subfigure}[b]{0.33\textwidth}
  \centering
  \includegraphics[width=5cm,height=6.5cm]{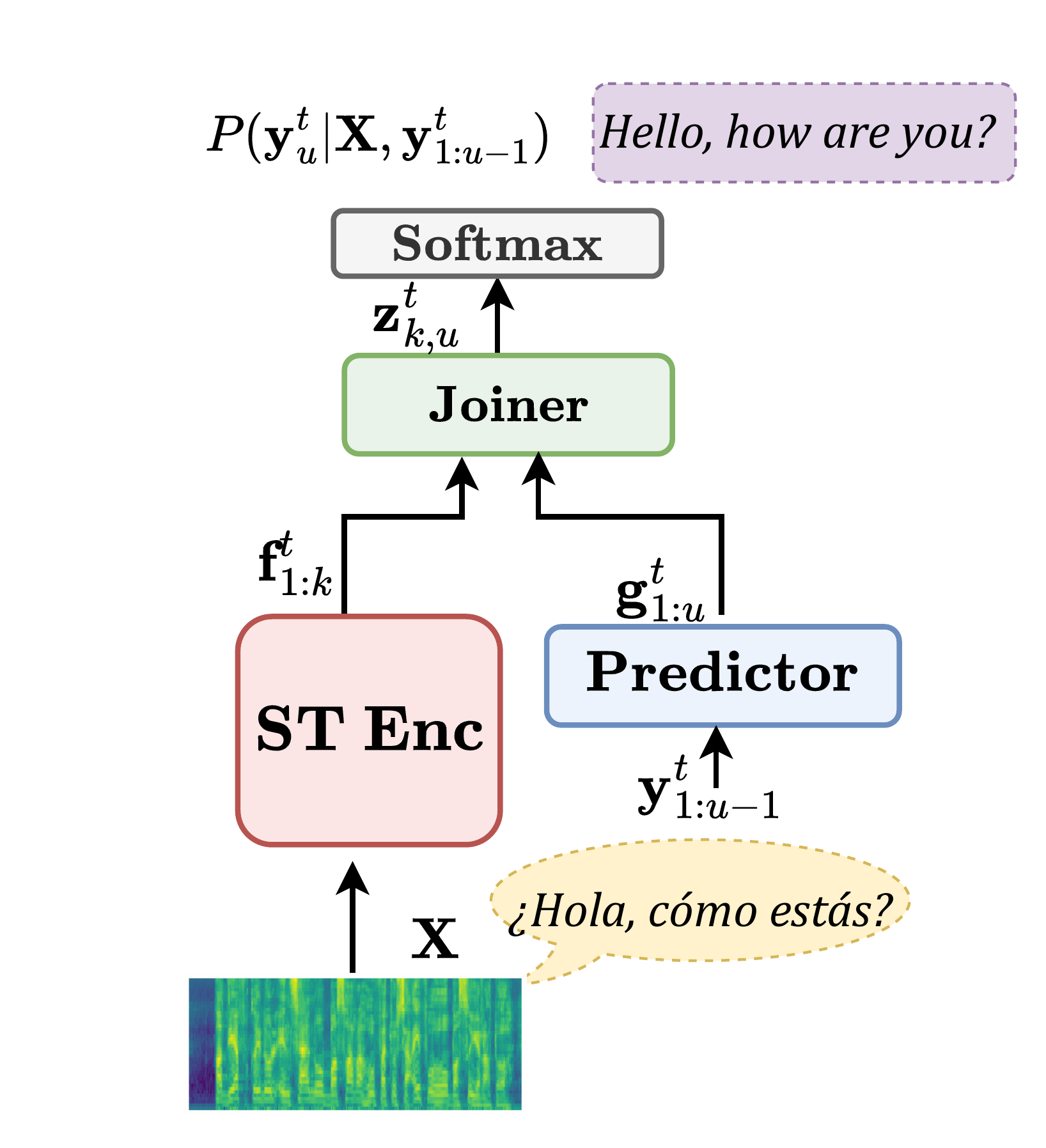}
  \vspace{+4mm}
  \caption{Neural Transducer based speech translation architecture}
  \label{fig:NT-ST}
  \end{subfigure}
  \hspace{5pt}
  \begin{subfigure}[b]{0.6\textwidth}
  \centering
  \includegraphics[width=12cm,height=9cm]{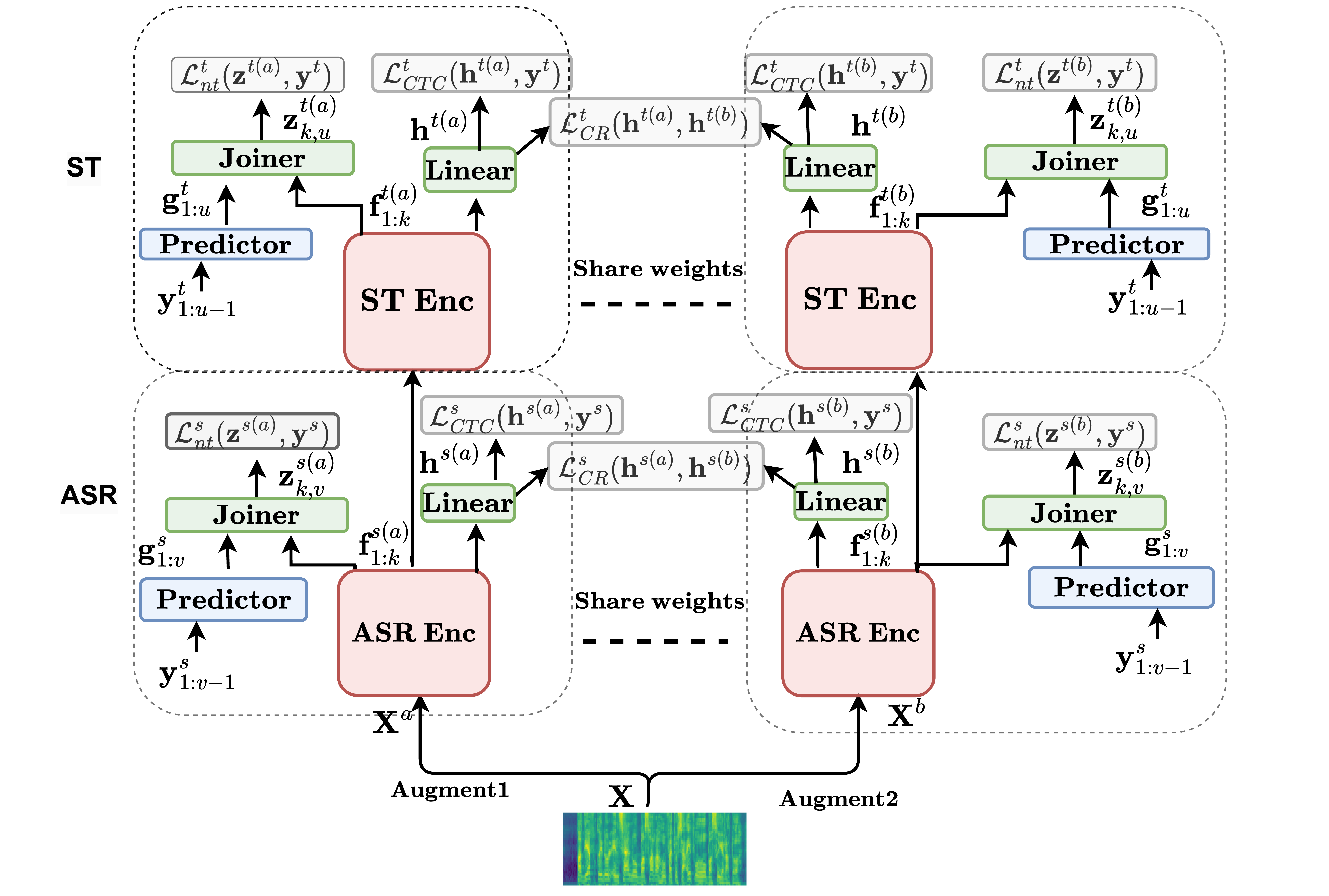}
  \vspace{-3mm}
  \caption{Proposed hierarchical neural transducer framework with self-distillation for joint speech recognition and translation (HENT-SRT).}
  \label{fig:HENT-SRT}
  \end{subfigure}
 \vspace{-2mm}
  \caption{Neural transducer-based speech translation and the proposed HENT-SRT architecture}
  \label{fig:proposed-method}
\end{figure*}

In this work, we propose a hierarchical neural transducer architecture, HENT-SRT,\footnote{Code is available at \url{https://github.com/k2-fsa/icefall}.} which decomposes speech translation (ST) into an automatic speech recognition (ASR) task followed by a translation task, effectively handling word reordering. The system employs multi-task training to optimize ASR and ST objectives simultaneously with a separate predictor and joiner for each task. Our objectives are three-fold: (1) to close the performance gap with state-of-the-art AED models in offline ST settings, (2) to improve computational efficiency during training and inference, and (3) to jointly model ASR and ST while maintaining ASR performance. To this end, we introduce HENT-SRT, a neural transducer-based ST framework that addresses (1) through a hierarchical encoder architecture, enabling more effective handling of reordering in translation. To improve ST efficiency in (2), we integrate a stateless 1D-CNN predictor~\citep{ghodsi2020rnn} and adopt the Zipformer architecture~\citep{yao2024zipformer}, which achieves superior ASR performance compared to state-of-the-art AED models such as E-Branchformer~\citep{kim2023branchformer}, while offering greater computational efficiency. Additionally, to reduce training complexity, we employ the pruned transducer loss~\citep{Kuang2022PrunedRF}. To mitigate ASR degradation in joint modeling for (3), we employ self-distillation with CTC consistency regularization \citep{yao2024cr}. We evaluate the effectiveness of our approach on conversational speech translation across three language pairs: Tunisian Arabic-English, Spanish-English, and Chinese-English described in \citep{hussein2024enhancing}. Furthermore, we conduct a comprehensive ablation study to analyze the impact of each ST design choice in both offline and streaming setups.

\section{Proposed Approach}
Let $\mathbf{X} = (\mathbf{x}_1, \ldots, \mathbf{x}_T) \in \mathbb{R}^{T \times F}$ denote the source speech, a sequence of $T$ acoustic feature vectors of dimension $F$. The goal is to generate a target word sequence $\mathbf{y}^{(t)} = (y^{(t)}_1, \ldots, y^{(t)}_U) \in \mathcal{V}^{(t)}_U$, a translation of length $U$. We use superscripts $(s)$ and $(t)$ to refer to the \emph{source} and \emph{target} languages, respectively. Speech translation task is trained using discriminative learning by minimizing the negative log-likelihood $\mathcal{L} = -\log P(\mathbf{y}^{(t)}|\mathbf{X})$. Transducers compute this probability by marginalizing over the set of all possible alignments $\mathbf{a} \in \bar{\mathcal{V}}^{(t)}_{T+U}$ as follows:
\begin{equation}
P(\mathbf{y}^{(t)}|\mathbf{X}) = \sum_{\mathbf{a}\in \mathcal{B}^{-1}(\mathbf{y}^{(t)})} P(\mathbf{a}|\mathbf{X})
\label{eq:1}
\end{equation}
where $\bar{\mathcal{V}}^{(t)} = \mathcal{V}^{(t)}\cup \{\phi\}$, $\phi$ is a {\em blank} label and $\mathcal{B}:\bar{\mathcal{V}}^{(t)}_{T+U}\to\mathcal{V}^{(t)}_U$ is the deterministic mapping from an alignment $\mathbf{a}$ to the sequence $\mathbf{Y}^{(t)}$ of its non-blank symbols. Transducers parameterize $P(\mathbf{a}|\mathbf{X})$ using an encoder, a prediction network, and a joiner, as illustrated in Figure~\ref{fig:NT-ST}. The encoder maps $\mathbf{X}$ to a representation sequence $\mathbf{f}^{\text{t}}_{1:k}, k\in\{1,\ldots,T\}$,
the predictor transforms $\mathbf{y}^{(t)}$ sequentially into $\mathbf{g}_{1:u}^{\text{t}}, u\in\{1,\ldots,U\}$, and the joiner combines $\mathbf{f}^{\text{t}}_{1:k}$ and $\mathbf{g}^{\text{t}}_{1:u}$ to generate logits $z^{t}_{k,u}$ whose softmax is the posterior distribution of $a_k$ (over $\bar{\mathcal{V}}^{(t)}$), i.e.
\begin{equation}
    \begin{aligned}
    P(\mathbf{y}^{(t)}|\mathbf{X}) &= \sum_{\mathbf{a} \in \mathcal{B}^{-1}(\mathbf{y}^{(t)})} \prod_{k=1}^{T+U} P(\mathbf{a}^{(t)}_k|f^{t}_{1:k},g^{t}_{1:u(k)}) \\
    &= \sum_{\mathbf{a} \in \mathcal{B}^{-1}(\mathbf{y}^{(t)})} \prod_{k=1}^{T+U} \text{softmax}(z^{st}_{k,u(k)})
    \end{aligned}
    \label{eq:2}
\end{equation}
where $u(k)\in  \{1, \cdots ,U \}$ denotes the index in the label sequence at time $k$. The negative log of the quantity in (\ref{eq:2}) is known as the transducer loss.
%

\subsection{ST hierarchical encoder}
To enhance the model’s ability to handle word reordering during translation, we propose a hierarchical architecture by adding a translation-specific encoder on top of the ASR encoder, as illustrated in Figure~\ref{fig:HENT-SRT}. This design enables joint modeling of ASR and ST while decomposing ST into ASR and translation tasks. Unlike~\citet{wang23oa_interspeech}, our translation-specific encoder facilitates more flexible reordering over the latent, monotonic ASR representations. Inspired by~\citet{dalmia2021searchable}, we adopt a two-stage training strategy: (1) ASR pretraining, followed by (2) multitask fine-tuning with both ST and ASR objectives. The ASR encoder, $\mathrm{ENC_{asr}}(\cdot)$, maps the input acoustic features $\mathbf{X}$ to a latent representation, $\mathbf{f}^{\mathrm{s}}$, as defined in Eq.~(\ref{eq3}).
\begin{equation}
    \mathbf{f}^{\mathrm{s}} = ENC_{\mathrm{asr}}(\mathbf{X})
    \label{eq3}
\end{equation}
Following this, the representation, $\mathbf{f}^{\mathrm{s}}$, serves as input to the ST encoder, $\mathrm{ENC^{st}}(\cdot)$, demonstrated in Eq. (\ref{eq4}) 
\begin{equation}
    \mathbf{f}^{\mathrm{t}} = ENC_{\mathrm{st}}(\mathbf{f}^{\mathrm{s}})
    \label{eq4}
\end{equation}
The transducer loss is computed for both ASR ($\mathcal{L}^s_{nt}$) and ST ($\mathcal{L}^t_{nt}$) objectives following Eq.~(\ref{eq:2}). The overall multitask objective, $\mathcal{L}_{nt}$, is the weighted sum of the two objectives:
\begin{equation}
    \mathcal{L}_{nt} = \alpha_{asr}\mathcal{L}^s_{nt} + \alpha_{st}\mathcal{L}^t_{nt}
    \label{eq5}
\end{equation}
where $\alpha_{asr}$ and $\alpha_{st}$ are hyperparameters controlling the contribution of ASR and ST losses, respectively.

\subsection{CR-CTC self-distillation}
Balancing multitask ASR and ST optimization in a two-stage training approach is challenging, as it often improves ST performance at the cost of ASR degradation. To achieve robust ST performance with minimal ASR degradation, we employ self-distillation with consistency-regularized CTC (CR-CTC)~\cite{yao2024cr}. This approach applies SpecAugment~\cite{park2019specaugment} to generate two augmented views of the same input, $\mathbf{X}^a$ and $\mathbf{X}^b$, which are then processed by a shared ASR encoder:  
\begin{align}
    \mathbf{f}^{s(a)} &= \mathrm{ENC}_{\mathrm{asr}}(\mathbf{X}^{(a)}) \\
    \mathbf{f}^{s(b)} &= \mathrm{ENC}_{\mathrm{asr}}(\mathbf{X}^{(b)})
\end{align}
The CR-CTC framework optimizes two objectives: the CTC loss $\mathcal{L}_{CTC}$ and the consistency regularization $\mathcal{L}_{CR}$, computed using the Kullback-Leibler divergence ($D_{KL}$) between encoder outputs:
\begin{align}
    \mathcal{L}_{CTC} &= \frac{1}{2} (\mathcal{L}_{CTC}(\mathbf{h}^{(a)}, \mathbf{y}) \nonumber\\
    &\quad + \mathcal{L}_{CTC}(\mathbf{h}^{(b)}, \mathbf{y})) \label{eq6}
\end{align}
\begin{align}
    \mathcal{L}_{CR} &= \frac{1}{2} \sum_{k=1}^{T} 
    \Big( D_{KL}(\text{sg}(h_k^{(b)}) \parallel h_k^{(a)})  \nonumber \\
    &\quad + D_{KL}(\text{sg}(h_k^{(a)}) \parallel h_k^{(b)}) \Big)
\end{align}
where $\text{sg}(\cdot)$ denotes the stop-gradient operation. In the proposed HENT-SRT framework, $\mathcal{L}_{CR}$ and $\mathcal{L}_{CTC}$ are computed for both ASR and ST encoders, as illustrated in Figure~\ref{fig:HENT-SRT}. The final objective is optimized using a multitask learning formulation that combines CR and CTC losses from both ASR and ST tasks along with the NT loss from Eq.~\ref{eq5}:
\begin{align}\label{eq:7}
    \mathcal{L} &= \mathcal{L}_{\text{nt}} 
    + \alpha_{\text{CR(asr)}} \mathcal{L}^s_{\text{CR}}
    + \alpha_{\text{CTC(asr)}} \mathcal{L}^s_{\text{CTC}} \nonumber \\
    &\quad + \alpha_{\text{CR(st)}} \mathcal{L}^t_{\text{CR}} + \alpha_{\text{CTC(st)}} \mathcal{L}^t_{\text{CTC}}
\end{align}
where $\alpha$ values are hyperparameters controlling the relative contributions of each loss term.

\subsection{ST decoding}\label{sec:decoding}
The translation decoding objective is to find the most probable target sequence $\hat{\mathbf{y}}^{(t)}$ among all possible outputs $\mathbf{y}^{(t)*}$ by maximizing the log-likelihood:
\begin{equation}
    \hat{\mathbf{y}}^{(t)} 
    = \argmax_{\mathbf{y}^{(t)*}} \sum_{l=1}^{U} \log P(\mathbf{y}^{(t)}_l | \mathbf{y}^{(t)}_{<l}, \mathbf{X})
\end{equation}
Neural transducers can model word reordering by using blank emissions to delay output tokens and then emitting multiple tokens within a single time step, as shown in Figure~\ref{fig:transducer}. 
 \begin{figure}[tb!]
\centering
\includegraphics[width=0.9\columnwidth]{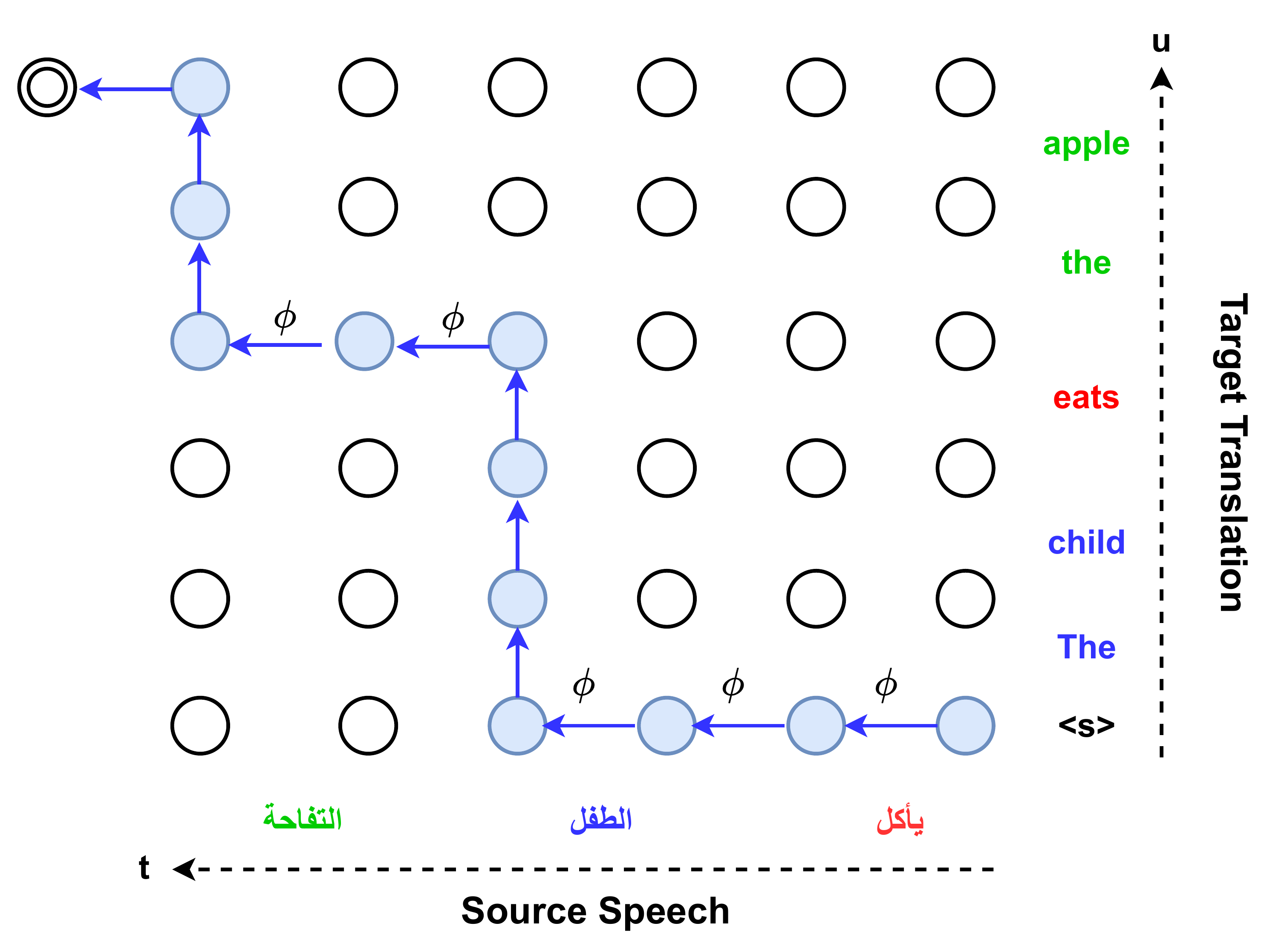}
\caption{Illustration of the transducer decoding graph for an ST task.}
\label{fig:transducer} 
\end{figure}
One challenge with transducer decoding is that the input audio features are typically longer than the output sequence, non-spoken regions are mainly represented by blank tokens, leading to a bias toward blank emissions~\cite{mahadeokar2021alignment}. This bias increases deletions and degrades translation quality. To mitigate this issue and control blank emissions during decoding, we introduce a blank penalty (BP) by adjusting the logit scores $\mathbf{z}$ in the log space:
\begin{equation} 
    z^t[:, 0] = z^t[:, 0] - BP
\end{equation}

\subsection{Computation efficiency}
To improve computational efficiency during training and inference, we adopt best practices from ASR transducer and examine their impact on ST in Section~\ref{sec:ablation}. For the encoder, we utilize the recently proposed Zipformer~\citep{yao2024zipformer}, which consists of multiple encoder blocks operating at different down-sampling rates. This hierarchical structure reduces the number of frames to process, thereby lowering computational complexity. Additionally, we replace the traditional LSTM-based prediction network with a stateless 1D-CNN trigram model~\citep{ghodsi2020rnn}.
Another major drawback of NT is the high memory consumption of the transducer loss due to the need for marginalization over logit tensor of size (batch, T, U, Vocab). To address this, we replace the full-sum transducer loss with the pruned transducer loss, which applies a simple linear joiner to prune the alignment lattice before computing the full sum on the reduced lattice~\citep{Kuang2022PrunedRF}. This significantly reduces memory overhead while maintaining competitive performance.

\section{Experiments}
\subsection{Experimental setup}\label{sec:expsetup}
We evaluate the effectiveness of our proposed approach on three conversational datasets: Fisher Spanish-English \citep{post2013improved}, Tunisian Arabic-English \citep{antonios2022findings}, and HKUST Chinese-English \citep{wotherspoon2024advancing}.
These datasets contain $3$-way data comprising telephone speech, source language transcriptions, and corresponding English translations.  \\ 
\textbf{Data pre-processing:} Our experiments utilize the Icefall framework,\footnote{\url{https://github.com/k2-fsa/icefall}} with the Lhotse toolkit \citep{zelasko2021lhotse} for speech data preparation. 
All audio recordings are resampled from $8$kHz to $16$kHz and augmented with speed perturbations at factors of $0.9$, $1.0$, and $1.1$. We extract 80-dimensional mel-spectrogram features using a 25ms window with a 10ms frame shift. 
Additionally, we apply on-the-fly SpecAugment \citep{park2019specaugment}, incorporating time warping (maximum factor of 80), frequency masking (two regions, max width of 27 bins), and time masking (ten regions, max width of 100 frames). For CR-CTC self-distillation, we follow \citep{yao2024cr} and increase both the number of time-masked regions and the maximum masking fraction by a factor of 2.5. For vocabulary, we employ a shared Byte Pair Encoding (BPE) vocabulary of size $5000$ for multilingual source transcriptions and $4000$ for English target translations.
 \\
\textbf{Models:} As a baseline, we reproduce the multitask (ASR+ST) neural transducer with a shared encoder \citep{wang23oa_interspeech}, employing the Zipformer architecture \citep{yao2024zipformer}. The model consists of $8$ blocks, each containing $2$ self-attention layers. The number of attention heads per block is set to \{4, 4, 4, 8, 8, 8, 4, 4\}, with attention dimensions of \{192, 256, 384, 512, 512, 384, 384, 256\}. The feed-forward dimensions per block are configured as \{512, 768, 1024, 1024, 1024, 1024, 1024, 768\}, and the convolution kernel sizes are \{31, 31, 15, 15, 15, 15, 31, 31\}. For the proposed hierarchical approach, we explore two architectural variants: \textit{shallow} (NT-Hier1) and \textit{deep} (NT-Hier2). In the \textit{shallow} configuration, we partition the baseline encoder by assigning the first five blocks to the ASR encoder $\mathrm{ENC_{asr}}$ in Eq.~(\ref{eq3}) and the remaining three blocks to the ST encoder $\mathrm{ENC_{st}}$ in Eq.~(\ref{eq4}). In contrast, the \textit{deep} variant increases the depth of the ST encoder to five blocks while reducing the number of parameters per block by approximately half, thereby maintaining the overall model size. Both the baseline and hierarchical variants contain approximately $70$M parameters. For the ablation studies in Sec.~\ref{sec:ablation}, which focus on a single language pair, we use a smaller model by reducing the depth of both the ASR and ST encoders by one block, resulting in a total of 45M parameters. We experiment with two types of predictor networks: a stateless predictor implemented as a single Conv1D layer with kernel size $2$, and a stateful LSTM predictor, both with hidden size $256$. Our training configuration utilizes the ScaledAdam optimizer \citep{yao2024zipformer} with a learning rate of $0.045$. 
The multitask weights $\alpha_{asr}$ and $\alpha_{st}$ in Eq.~(\ref{eq5}) are set to $1$, while $\alpha_{CR(asr)}$ and $\alpha_{CR(st)}$ are set to $0.05$, and $\alpha_{CTC(asr)}$ and $\alpha_{CTC(st)}$ are set to $0.1$ in Eq.~(\ref{eq:7}). For comparison, we use the ESPnet CTC/Attention ST model\footnote{\url{https://github.com/espnet/espnet/blob/master/egs2/must_c_v2/st1/conf/tuning/train_st_ctc_conformer_asrinit_v2.yaml}} with 75M parameters. We choose the Conformer encoder, as it outperforms eBranchformer in translation quality.
The optimal learning rate and warmup steps, set to $0.001$ and $30$k, are selected from the ranges \{0.0005–0.002\} and \{15k–40k\}, respectively. All models are trained for $30$ epochs on $4$ V$100$ GPUs, using a batch size of $400$ seconds. Unless otherwise noted, all decoding results reported in the offline setting (Sec.~\ref{soa}) use beam search with a beam size of 20.
 \\
\textbf{Evaluation:} To ensure a comprehensive evaluation of translation quality, we utilize BLEU \citep{papineni2002bleu} for surface-level word matching, chrF++ for character-level accuracy, and COMET \citep{rei2020comet}\footnote{We used \texttt{Unbabel/wmt22-comet-da} model.} for semantic adequacy. Translation evaluation is conducted in a case-insensitive manner, without punctuation. ASR performance is assessed using word error rate (WER). To assess the processing delay of the system during streaming, we report the real-time factor (RTF).

\begin{table}[tb!]
    \centering
     \caption{Ablation for transducer based ST, including blank-penalty (BP), lattice pruning-range, scaling warmup-steps, predictor and encoder structure.}
     \setlength{\tabcolsep}{2pt}
\resizebox{0.49\textwidth}{!}{
\begin{tabular}{l c c c c c c c c}
\toprule
\textbf{Model} &  & \textbf{Prune-} & \textbf{Warmup-} & \textbf{Pre-} 
& \multicolumn{2}{c}{\textbf{Dev1}} 
& \multicolumn{2}{c}{\textbf{Dev2}} \\
\cmidrule(lr){6-7} \cmidrule(lr){8-9}
& \textbf{BP} &\textbf{range} &\textbf{steps} & \textbf{dictor}& \textbf{WER $\downarrow$} & \textbf{BLEU $\uparrow$} & \textbf{WER $\downarrow$} & \textbf{BLEU $\uparrow$} \\
\midrule
NT (ASR) & 0 & 5 & 20k & CNN & \textbf{40.0} & - & \textbf{42.4} & - \\
NT (ST)  & 0 & 5 & 20k & CNN & - & 14.7 & - & 12.4 \\
NT (ST)  & 0 & 10 & 20k & CNN & - & 16.2 & - & 13.2 \\
NT (ST)  & 0 & 10 & 20k & LSTM & - & 15.8 & - & 13.0 \\
NT (ST)  & 0 & 10 & 5k & CNN & - & 18.1 & - & 14.7 \\
\midrule
NT-Hier1 (ASR+ST)  & 0 & 10 & 5k & CNN & 40.0 & 18.5 & 42.8 & 15.7 \\
NT-Hier2 (ASR+ST)  & 0 & 10 & 5k & CNN & 40.3 & \textbf{19.0} & 43.6 & \textbf{15.8} \\
\midrule
NT-Hier2 (ASR+ST)  &1 & 10 & 5k & CNN & 40.3 & \textbf{20.4} & 43.6 & \textbf{17.1} \\
\bottomrule
\end{tabular}
}

 \vspace{-0.3cm}
    \label{tab:ablation}
\end{table}

\begin{table*}[tb!]
\centering
\caption{Comparison of ASR and ST performance between the state-of-the-art Neural Transducer (NT) with a shared encoder and a multi-decoder AED (MultiDec). ASR performance is measured using WER, while ST performance is evaluated using BLEU, chrF++, and COMET.} 
\vspace{-5pt}
\resizebox{\textwidth}{!}
{%
\begin{tabular}{lccccccccccccc}
\toprule
\textbf{Model} &  & \multicolumn{4}{c}{\textbf{Tunisian}} & \multicolumn{4}{c}{\textbf{HKUST-Chinese}} & \multicolumn{4}{c}{\textbf{Fisher-Spanish}} \\
\cmidrule(lr){3-6} \cmidrule(lr){7-10} \cmidrule(lr){11-14}
& \textbf{BP} & \textbf{WER $\downarrow$} & \textbf{BLEU $\uparrow$} & \textbf{chrF++ $\uparrow$} & \textbf{COMET $\uparrow$} 
& \textbf{WER $\downarrow$} & \textbf{BLEU $\uparrow$} & \textbf{chrF++ $\uparrow$} & \textbf{COMET $\uparrow$} 
& \textbf{WER $\downarrow$} & \textbf{BLEU $\uparrow$} & \textbf{chrF++ $\uparrow$} & \textbf{COMET $\uparrow$} \\
\midrule
NT (ASR)  &-& \textbf{41.4} & - & - & - & \textbf{22.8} & - & - & -  & 18.2 & - & - &  \\
NT (ST)  &-& - & 15.3 & 35.9 & 0.656 & - & 10.0 & 30.5 & 0.711  & - & 30.6 & 56.0 & 0.793 \\

NT-Shared (ASR+ST) \cite{wang23oa_interspeech}  & - & 41.6 &16.3 & 37.1 & 0.660 & 23.8 & 10.4 & 30.9 & 0.714 & \textbf{18.0}& 31.0 & 56.4 & 0.798 \\
NT-Hier1 (ASR+ST)  &-& 42.6 & 17.8 & 40.4 & 0.670 & 23.5 & 11.3 & 33.8 & 0.720 & 18.3 & 31.9 & 58.2 & 0.801\\
NT-Hier2 (ASR+ST)  &-& 43.1 & 18.3 & 40.6 & 0.672 & 23.9 & 12.0 & 33.9 &  0.722 & 18.9 & 32.4  & 58.5 & \textbf{0.801} \\
NT-Hier2 (ASR+ST)  & 0.5 & 43.1 & \textbf{19.4} & \textbf{42.6} & \textbf{0.674} & 23.9 & \textbf{12.9} & \textbf{36.2} &  \textbf{0.724} & 18.9 & \textbf{33.0}  & \textbf{59.9} & \textbf{0.801} \\
\midrule

CR-CTC (ASR) & - & \textbf{40.1} & - & - & - & \textbf{21.7} & - & -& -& \textbf{17.3} & - & - & - \\
HENT-SRT  & - & 41.4 & 17.8 & 39.8 & 0.675 &  22.8 & 11.5 & 32.7 &  0.726 &17.8  & 31.8 & 57.5 & \textbf{0.803} \\
HENT-SRT  & 1.0 & 41.4 & \textbf{20.6} & \textbf{43.4} & \textbf{0.682} &  22.8 & \textbf{14.7} & \textbf{37.5} &  \textbf{0.734} & 17.8  & \textbf{33.7} & \textbf{60.5} & \textbf{0.803} \\
\midrule
CTC/Attention (CA) \cite{yan2023espnet} &-& 42.7 & 20.4 & 44.2 & 0.680 & 24.6 & 15.2 & 38.8 & 0.706 & 18.9 & 33.9 & 60.8 & 0.796 \\

\bottomrule
\end{tabular}%
}
\vspace{-0.3cm}
\label{tab:soa}
\end{table*}

\subsection{Ablation analysis}\label{sec:ablation}
To assess the impact of efficiency-related design choices in our proposed hierarchical ST approach, we conduct ablation studies on the \textit{Tunisian-English} dataset, which includes two development sets for hyperparameter tuning, as shown in Table~\ref{tab:ablation}. To ensure fair comparisons, we maintain a consistent model size of approximately 45M parameters across all experiments. We begin by evaluating the vanilla pruned transducer (NT) architecture for ST, examining the effect of increasing the pruning range beyond the default value of $5$ (used in NT for ASR). Expanding the pruning range to $10$ leads to a BLEU improvement of up to $+1.5$, indicating that a larger alignment lattice enhances the model’s ability to perform word reordering during translation. However, further increases in pruning range provide only marginal additional gains.  Next, we explore the role of the prediction network in ST performance. Our results show that a simple trigram 1D-CNN stateless predictor performs comparably to a more complex LSTM-based predictor, consistent with previous observations in ASR~\citep{ghodsi2020rnn}. We also vary the number of training steps before fully incorporating the pruned loss, effectively controlling the balance between the simple and pruned objectives during early training. Reducing the number of warm-up steps yields BLEU gains of up to $+1.8$, suggesting that earlier exposure to pruning improves optimization. To assess the effect of hierarchical modeling, we compare shallow (NT-Hier$1$) and deep (NT-Hier$2$) ST encoders. To keep the parameter count constant, NT-Hier$2$ doubles the number of layers while halving the parameters per layer in the ST encoder. Our findings indicate that deeper architectures can improve ST performance by up to $+0.5$ BLEU, though at the cost of reduced ASR accuracy highlighting a trade-off between ST and ASR tasks. Finally, we evaluate the impact of a blank-penalty term applied during decoding. Introducing a small blank penalty leads to BLEU improvements of up to $+1.4$, demonstrating its effectiveness in refining translation quality. A more detailed analysis of the blank penalty is presented in Sec.~\ref{sec:streaming}.

\subsection{Comparison with state-of-the-art models}\label{soa}
In this section, we compare the proposed hierarchical neural transducer with self-distillation (HENT-SRT) model to state-of-the-art systems in the offline setting. Specifically, we evaluate against the multitask ASR-ST transducer with a shared encoder NT-Shared)~\cite{wang23oa_interspeech}, and the CTC/Attention-based AED model (CA)~\cite{yan-etal-2023-espnet}, shown in Table~\ref{tab:soa}.  Comparing the vanilla transducer models (NT-ASR and NT-ST) to NT-Shared, we find that a single multilingual ASR-ST model can be trained with the same number of parameters while maintaining comparable performance achieving up to +$1$ BLEU improvement in ST with at most +$1$ WER degradation in ASR. The hierarchical ST transducer models (NT-Hier) consistently outperform NT-Shared across all ST metrics, yielding improvements of up to +$2$ BLEU, +$3.5$ chrF++, and +$0.01$ COMET, but at the cost of up to +$2$ WER degradation in ASR performance. Furthermore, NT-Hier2, which employs a deeper ST encoder, achieves better translation performance up to +$0.7$ BLEU improvement over NT-Hier1 but incurs an additional ASR degradation of up to +$0.5$ WER. These findings suggest that the hierarchical two-stage training enhances translation quality, particularly in handling reordering (see Sec.~\ref{appendix}), but at the expense of ASR accuracy.
\begin{figure}[tb!]
\centering
\vspace{-0.4cm}
\includegraphics[width=1\columnwidth]{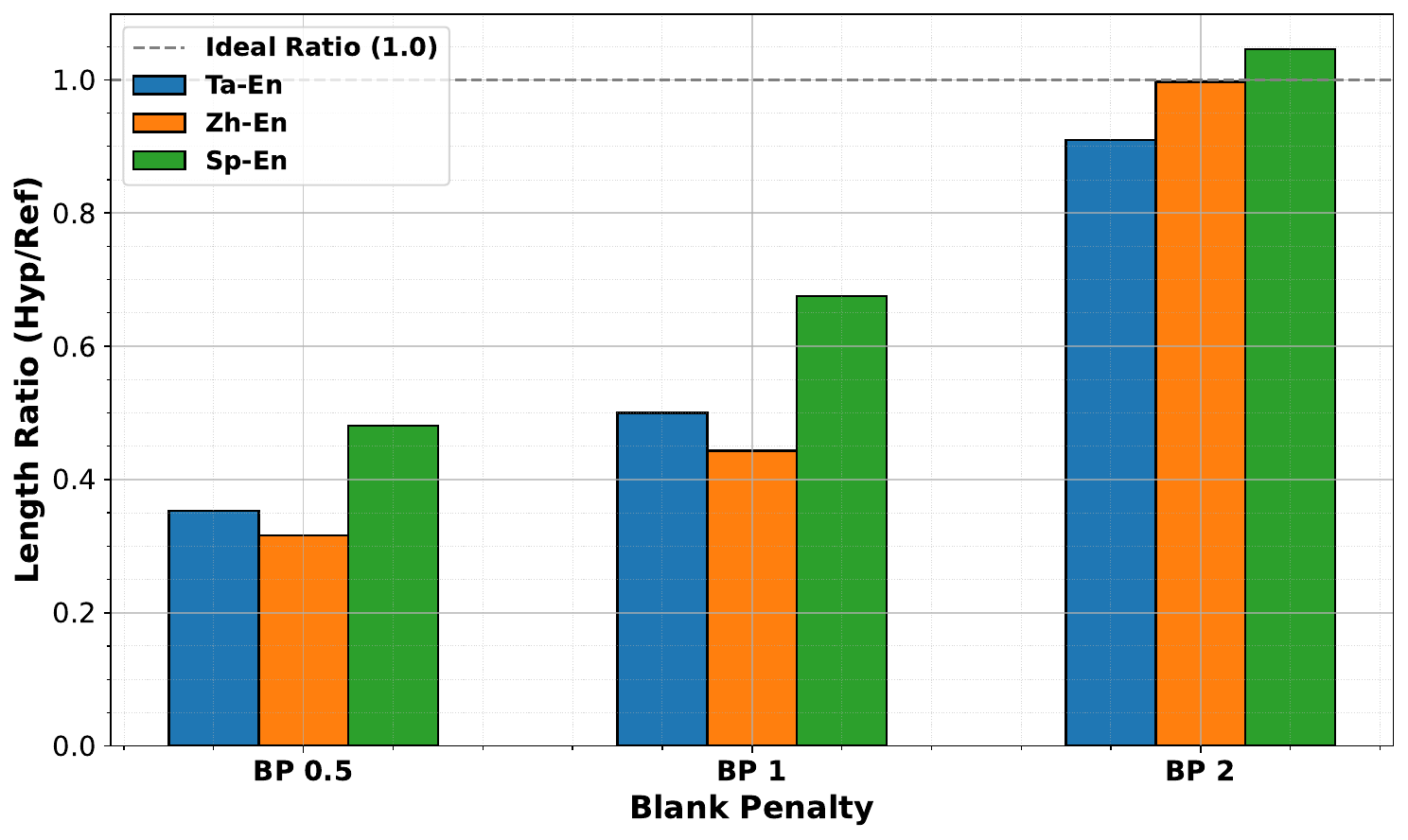}
\caption{Impact of the blank penalty on translation length ratio (hypothesis/reference) during greedy decoding for Tunisian (Ta-En), HKUST (Zh-En), and Fisher-Spanish (Sp-En). A ratio close to 1.0 indicates ideal length matching between hypothesis and reference.}
\vspace{-0.4cm}
\label{fig:blank_penalty} 
\end{figure} 
ST performance can be further improved by applying a blank penalty during decoding, yielding up to +$1$ BLEU and +$2.3$ chrF++ gains. To ensure robust ST performance with minimal ASR degradation, we integrate self-distillation with consistency-regularized CR-CTC to NT-Hier2, resulting in the proposed HENT-SRT model. Notably, HENT-SRT performs slightly worse than NT-Hier2 in ST when no blank penalty is applied. However, applying the optimal blank penalty to both models yields further gains for HENT-SRT, with improvements of up to +1.8 BLEU, +1.3 chrF++, and +0.01 COMET over NT-Hier2. Additionally, CR-CTC loss also acts as a regularizer, maintaining ASR performance close to the best results observed with CR-CTC(ASR). Finally, compared to the AED-based CA model~\cite{yan-etal-2023-espnet}, HENT-SRT closes the ST performance gap while achieving superior ASR performance.

\subsection{Streaming Translation}\label{sec:streaming}
In this section, we explore streaming speech translation using our proposed hierarchical ST framework. Streaming is simulated with greedy decoding by enabling causal convolution and applying chunked processing, using a chunk size of $64$ frames and a left context of $128$ frames. Specifically, the input is segmented into fixed-size chunks, and each chunk attends only to a fixed number of preceding chunks, with all future frames masked. To ensure low-latency decoding, we restrict the number of emitted symbols per frame to $20$, as larger values offer minimal additional improvements. We then examine the effect of the blank penalty on translation quality, as shown in Table~\ref{tab:streaming}. While tuning is performed on development sets, we report test set results for completeness. Across all datasets, increasing the blank penalty for NT-Hier2 up to a value of 2 consistently improves BLEU scores by reducing deletion errors, supporting the hypothesis presented in Section~\ref{sec:decoding}. This trend is further confirmed in Figure~\ref{fig:blank_penalty}, which shows that higher blank penalties result in longer translations that more closely align with reference lengths, with a penalty of 2 achieving near-perfect length matching. Overall, the proposed HENT-SRT framework achieves the best streaming performance across datasets, with BLEU improvements of up to +4.5 points compared to NT-Hier2. Importantly, the hierarchical ST design introduces no additional latency in terms of real-time factor compared to the vanilla NT model. This is expected, as the hierarchical structure reuses the original encoder in a sequentially factorized ASR-ST configuration without increasing computational complexity.

\begin{table}[tb!]
    \centering
    \vspace{-0.4cm}
    \caption{Comparison of translation BLEU scores and processing delays measured by real-time factor (RTF)}
    \setlength{\tabcolsep}{2pt}
\resizebox{0.49\textwidth}{!}{
\begin{tabular}{l c c c c c c c c c c}
\toprule
\textbf{Model} & \textbf{BP} & \multicolumn{3}{c}{\textbf{Tunisian}} 
& \multicolumn{3}{c}{\textbf{HKUST}} & \multicolumn{3}{c}{\textbf{Fisher-Sp}} \\
\cmidrule(lr){3-5} \cmidrule(lr){6-8} \cmidrule(lr){9-11}
& & \textbf{Dev1} & \textbf{Test} & \textbf{RTF} & \textbf{Dev} & \textbf{Test} & \textbf{RTF} & \textbf{Dev} & \textbf{Test} & \textbf{RTF} \\
\midrule
NT & 0.0 & 6.8 & 6.1 & 0.013 & 3.1 & 3.4 & 0.016 & 14.2 & 15.6 & 0.013 \\
NT-Shared  & 0.0 & 7.0 & 6.0 & 0.010 & 3.3 & 3.7 & 0.014 & 14.8 & 16.3 & 0.010 \\
NT-Hier2  & 0.0 & 9.5 & 9.0 & 0.012 & 4.2 & 4.6 & 0.013 & 16.8 & 17.9 & 0.012 \\
\midrule
NT-Hier2  & 0.5 & 10.4 & 10.0 & 0.012 & 5.0 & 5.7 & 0.012 & 19.3 & 19.6 & 0.012 \\
NT-Hier2  & 1.0 & 14.1 & 13.0 & 0.012 & 7.2 & 7.4 &0.015  & 24.8 & 25.6 & 0.012 \\
NT-Hier2  & 2.0 & 16.8 & 14.0 & 0.014 & 6.7 & 7.5 & 0.018 & 24.9 & 26.3 & 0.013 \\
\midrule
HENT-SRT  & 2.0 & \textbf{18.6} & \textbf{17.2} & 0.013 & \textbf{10.2} & \textbf{11.2} & 0.017 & \textbf{29.2} & \textbf{30.8} & 0.013 \\
\bottomrule
\end{tabular}
}

    \vspace{-0.4cm}
    \label{tab:streaming}
\end{table}

\section{Conclusion}
In this paper, we proposed HENT-SRT, a novel hierarchical transducer architecture for joint speech recognition and translation (ST). The model factorizes the ST task into two stages, ASR followed by a translation, enabling more effective handling of word reordering. To improve computational efficiency, HENT-SRT design incorporates key transducer-based ASR practices, including a downsampled encoder, stateless predictor, and pruned transducer loss. To maintain robust ST performance without sacrificing ASR accuracy, we apply self-distillation with CTC consistency regularization. Additionally, we introduce a blank penalty mechanism during decoding, which effectively reduces deletion errors and enhances translation quality. Experimental results show that HENT-SRT significantly outperforms previous state-of-the-art transducer-based ST models and closes the gap with attention-based encoder-decoder architectures, while achieving superior ASR performance. Moreover, our approach offers substantial gains in streaming scenarios without introducing additional delays.

\section*{Limitations}
In this work, we focus on non-overlapping speech translation, translating multilingual source speech into English text. For future work, we aim to extend our hierarchical approach to handle overlapped speech while expanding support for additional target languages and broader translation directions. 

\bibliography{acl2025}

\clearpage
\appendix

\section{Linguistic Analysis of Speech Translation Performance}\label{appendix}

\begin{CJK}{UTF8}{gbsn}
\begin{table*}[!tb]
\centering
\footnotesize
\caption{Comparison of NT-Hier2 and NT-Shared System Outputs on the Chinese test set}
\label{tab:chinese_bbn_comparison}
\begin{tabular}{c p{3cm} p{4cm} p{3cm} p{3cm}}
\toprule
\textbf{Row} & \textbf{Source (ref\_src)} & \textbf{Reference (ref\_tgt)} & \textbf{NT-Hier2} & \textbf{NT-Shared} \\
\midrule
1 & 没有问题对对没关系 & that's alright if there's no problem it's fine & no problem that's right it doesn't matter & no problem that's right it doesn't matter \\
2 & 那谁给他办的绿卡谁给他办的 & who applied that green card for him who did it for him & then who did it for him then who green it & then who who do it \\
3 & 我八月中旬就去上班然后再回来再回来答辩就是然后然后十二月份应该就可以毕业了 & i would go to work at the mid august then i would come back again for my thesis defence that's it and then i should be able to graduate in december & i will go to work in and then come back again and then i mean in december i can graduate in october & in august i will go to work on and then come back to work again and then i mean after october \\
\bottomrule
\end{tabular}
\end{table*}
\end{CJK}

\begin{CJK}{UTF8}{gbsn}
\subsection{Chinese HKUST Test Set Analysis}

The Chinese HKUST test set contains conversational Mandarin translated into English. The source utterances include informal structures, frequent repetitions (e.g., ``对对'' / ``yes yes''), fillers (e.g., ``呃'' / ``uh''), and discourse markers (e.g., ``啊'', ``哎呀''). We compare the NT-Hier2 and NT-Shared system outputs against reference translations to analyze their differences and better understand how hierarchical modeling improves translation quality. Table~\ref{tab:chinese_bbn_comparison} presents representative examples comparing the NT-Hier2 and NT-Shared systems.

\subsubsection{Observations and Comparison:}
\begin{itemize}
    \item \textbf{Syntactic Structure:} In longer utterances (e.g., Row 3), NT-Hier2 retains more of the reference structure, including mentions of "thesis defence" and a full sequence of events. However, it introduces timeline inconsistencies such as "in December I can graduate in October." Despite this, metrics like BLEU and COMET may still reward NT-Hier2's output for preserving semantic content.
    
    \item \textbf{Handling Reordering:} In Row 2, the NT-Hier2 output more accurately preserves the source's repeated clause structure (e.g., “who did it for him”), while the NT-Shared output compresses the translation and omits part of the repeated question. This indicates that NT-Hier2 is better at modeling long-range reordering and maintaining structural fidelity.
    
    \item \textbf{Handling Informal Speech:} Row 1 illustrates a relatively simple informal sentence, where both systems produce fluent and semantically equivalent outputs. This supports our observation that informal utterances with low lexical variability are easier to translate. However, in general, fillers and discourse markers may be translated literally, rephrased, or omitted depending on context, which introduces challenges for exact-match metrics like BLEU.
\end{itemize}
\end{CJK}

\textbf{BLEU vs. COMET Discrepancy:}  
The Chinese test set yields a BLEU score of 12.9 and a COMET score of 0.72, in contrast to Fisher Spanish (BLEU = 33, COMET = 0.80) and IWSLT23 Tunisian (BLEU = 19.4, COMET = 0.67). We hypothesize that the low BLEU score for Chinese BBN stems from significant lexical variation in the system outputs. COMET, on the other hand, is sensitive to semantic similarity and rewards outputs that convey approximate meaning (e.g., “no problem” aligns semantically with “it’s fine”). This discrepancy is consistent with the conversational nature of the dataset, where multiple valid translations (e.g., “it’s fine” vs. “it doesn’t matter”) are possible, thereby reducing BLEU’s reliability due to its reliance on a single reference.\\
\textbf{Pattern and Theory:}  
A recurring pattern in the NT-Hier2 system’s output is its tendency to produce more lexically diverse and expressive translations. In contrast, the NT-Shared system often generates shorter, simpler outputs, suggesting possible underfitting or reduced modeling capacity. We argue that the low BLEU score reflects not only actual translation errors but also the limitations of BLEU in evaluating Chinese-to-English translation, especially in high-variability, conversational settings. Future work could explore multi-reference BLEU or fine-tuned semantic metrics to better capture these nuances.


\end{document}